\documentclass[10pt,twocolumn,letterpaper]{article}

\usepackage{cvpr}
\usepackage{times}
\usepackage{epsfig}
\usepackage{graphicx}
\usepackage{amsmath}
\usepackage{amssymb}
\usepackage{multirow}
\usepackage{balance} 
\usepackage{verbatim}

\usepackage[pagebackref=true,breaklinks=true,letterpaper=true,colorlinks,bookmarks=false]{hyperref}

\cvprfinalcopy 

\def\cvprPaperID{2611} 

\ifcvprfinal\fi
\begin{document}

\title{Fine-grained Image Classification via Combining Vision and Language}

\author{Xiangteng He and Yuxin Peng\thanks{Corresponding author.} \\
Institute of Computer Science and Technology, Peking University\\
{\tt\small hexiangteng@pku.edu.cn, pengyuxin@pku.edu.cn}
}

\maketitle
\thispagestyle{empty}

\begin{abstract}
Fine-grained image classification is a challenging task due to the \textbf{large} intra-class variance and \textbf{small} inter-class variance, aiming at recognizing hundreds of sub-categories belonging to the same basic-level category. Most existing fine-grained image classification methods generally learn part detection models to obtain the semantic parts for better classification accuracy. Despite achieving promising results, these methods mainly have two limitations: (1) not all the parts which obtained through the part detection models are beneficial and indispensable for classification, and (2) fine-grained image classification requires more detailed visual descriptions which could not be provided by the part locations or attribute annotations. For addressing the above two limitations, this paper proposes the two-stream model \textbf{combining vision and language (CVL)} for learning latent semantic representations. \textbf{The vision stream} learns deep representations from the original visual information via deep convolutional neural network. \textbf{The language stream} utilizes the natural language descriptions which could point out the discriminative parts or characteristics for each image, and provides a flexible and compact way of encoding the salient visual aspects for distinguishing sub-categories. Since the two streams are complementary, combining the two streams can further achieves better classification accuracy. Comparing with 12 state-of-the-art methods on the widely used CUB-200-2011 dataset for fine-grained image classification, the experimental results demonstrate our CVL approach achieves the best performance.
\end{abstract}

\section{Introduction}
\begin{figure}[t]
\begin{center}
\includegraphics[width=0.9\linewidth]{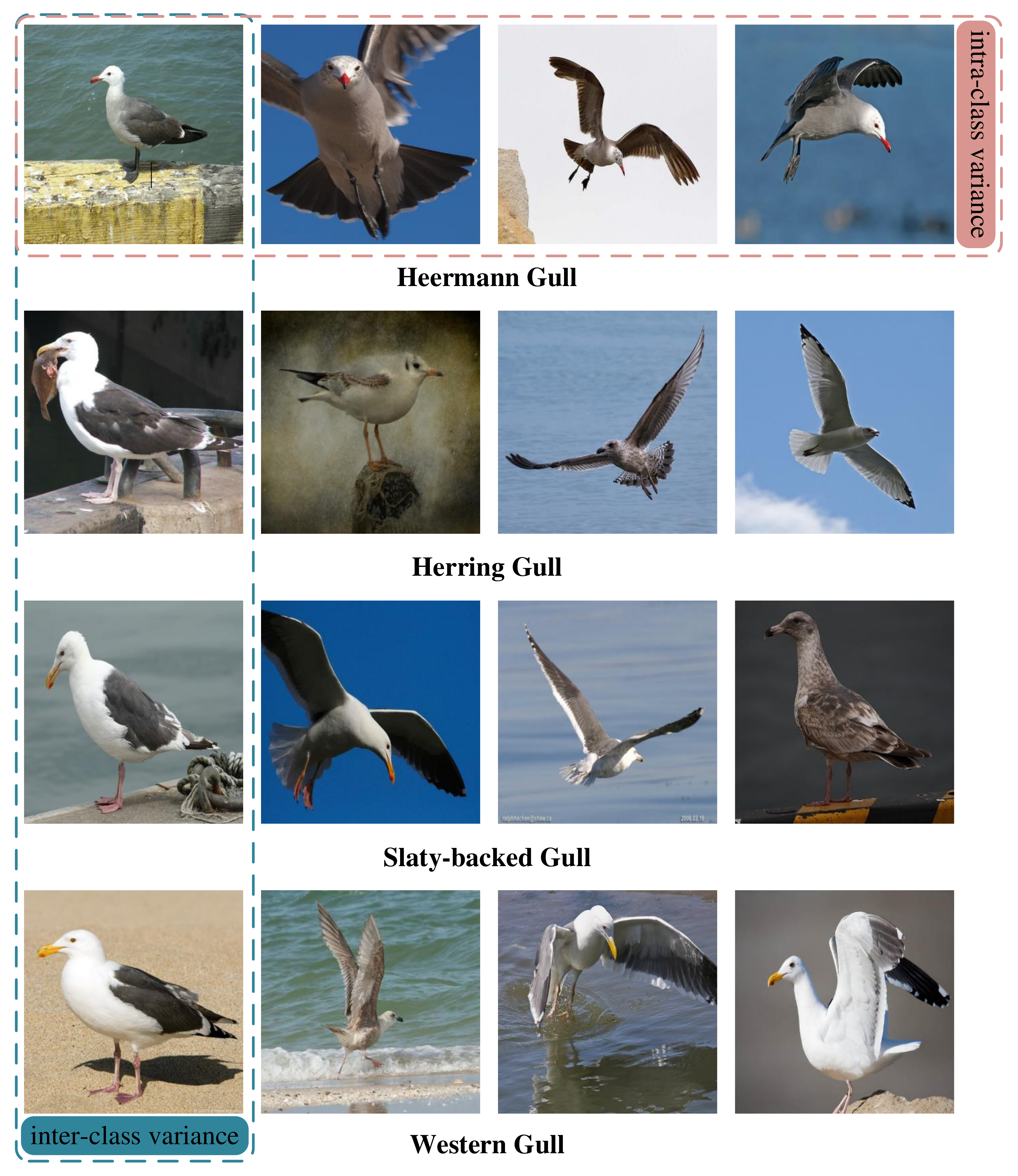}
\end{center}
   \caption{Examples from CUB-200-2011. Note that it is a technically challenging task even for humans to categorize them due to large intra-class variance and small inter-class variance.}
\label{example}
\end{figure}

Fine-grained image classification  aims to recognize sub-categories under some basic-level categories. Models of fine-grained image classification have made great progress in recent years\cite{partstacked,twoattention,partrcnn,spatialconstraints,picking}, due to the progress of deep neural networks. And on the data side, more fine-grained domains have been covered, such as bird types \cite{birdsnap,cub2011}, dog species \cite{stanforddog}, plant breeds \cite{plant} , car types \cite{krause20133d} and aircraft models \cite{aricraft}.It is easy for an inexperienced person to recognize basic-level categories such as bird, flower and car, but highly hard to recognize 200 or even more sub-categories. Consequently, fine-grained image classification is a technically challenging task, due to the large intra-class variance and small inter-class variance, as shown in Figure \ref{example}.

\par 
The sub-categories are generally same in global appearance, and distinguished by the subtle and local differences, such as the color of abdomen, the shape of toe and the texture of feather for bird. These subtle differences are located at the regions of object or its parts, so the localization of object and its parts is crucial for fine-grained image classification. A two-stage learning framework is adopted by most of the existing methods: the first stage is to localize the object or its discriminative parts, and the second stage is to extract the deep features of the object or its parts through Convolutional Neural Network (CNN) and train a classifier for the final prediction. However, these methods mainly have two limitations. First, parts are crucial for achieving better accuracy, but not all of them are beneficial and indispensable. Huang et al. \cite{partstacked} show that the classification accuracy declines when the number of parts increases from 8 to 15 in the experiments of their Part-stacked CNN method. Zhang et al. \cite{picking} pick only 6 parts in the experiment for achieving the best classification accuracy. And Zhang et al. \cite{partrcnn} only use the head and body parts for classification in Part-based R-CNN. Generally speaking, the number of the parts used in the experiments is highly empirical. This is highly limited in flexibility, and difficult for generalizing to the other datasets or domains. Second, fine-grained image classification requires more detailed visual descriptions which could not be provided by the part locations or attribute annotations. The part locations could not point out which part is the discriminative parts for sub-categories recognition and tell the discriminative features, such as the color of bill and the shape of wing. The attribute annotations may tell us the color of bill, but they do not tell us whether the bill is significantly important for distinguishing sub-categories than other parts. Hence, we need a kind of promising information to tell us the attributes as well as the importances of the parts, and further facilitate the classification accuracy. Fortunately, text descriptions from natural language satisfies the above promising requirements.
\par
How to exactly relate the natural language descriptions to the visual content of images is the key of image classification. Inspired by the progress of the cross-modal analysis which reduces the multi-modal representation gap between visual information and natural language descriptions, this paper proposes a two-stream model combining vision and language (CVL) for learning latent semantic representations. The vision stream first localizes the object of image via saliency extraction and co-segmentation, and then learns deep representations of the original image and its discriminative object via deep convolutional neural network. The language stream utilizes the cross-modal analysis to learn the correlation between the natural language descriptions and the discriminative parts, and provides a flexible and compact way of encoding the salient visual aspects for distinguishing sub-categories. Vision stream focuses on the locations of the discriminative regions, while language stream focuses on the attributes of the discriminative regions. They are complementary, combining the two streams further exploit the correlation between visual feature and nature language descriptions, and enhances their mutual promotion to achieve better classification accuracy. Comparing with 12 state-of-the-art methods on the widely used CUB-200-2011 dataset for fine-grained image classification, the experimental results demonstrate our CVL approach achieves the best performance.
\par
The rest of this paper is organized as follows: Section \uppercase\expandafter{\romannumeral2} briefly reviews related works on fine-grained image classification and cross-modal analysis. Section \uppercase\expandafter{\romannumeral3} presents our proposed CVL approach, and Section \uppercase\expandafter{\romannumeral4} introduces the experiments as well as the results analyses. Finally Section \uppercase\expandafter{\romannumeral5} concludes this paper.

\section{Related Work}
\subsection{Fine-grained Image Classification}
Most existing works follow the pipeline: first localizing the object or its parts, and then extracting discriminative features for fine-grained image classification. Some fine-grained image classification datasets, e.g. CUB-200-2011 \cite{cub2011}, have the detailed annotations: object annotation (i.e. bounding box of object) and parts annotations (i.e. parts locations), an intuitive idea is that using these annotations for the localizations of object and its parts. Object annotation is used in the works of \cite{chai2013symbiotic,yang2012unsupervised} to learn part detectors in an unsupervised or latent manner. And even part annotations are used in these methods \cite{berg2013poof,xie2013hierarchical}. Since the annotations of the testing image are not available in practical applications, some researchers use the object or part annotations only at training stage and no knowledges of any annotations at testing stage. Object and Part annotations are directly used in training phase to learn a strongly supervised deformable part-based model \cite{zhang2013deformable} or directly used to fine-tune the pre-trained CNN model \cite{branson2014bird}. Further more, Krause et al. \cite{krause2015fine} only use object annotation at training stage to learn the part detectors, then localize the parts automatically in the testing stage. Recently, there are some promising works attempting to learn the part detectors under the weakly supervised condition, which means that neither object nor part annotations are used in both training and testing phase. These works make it possible to put the fine-grained image classification into practical applications. Simon et al. \cite{simon2015neural} propose a neural activation constellations part model (NAC) to localize parts with constellation model. Xiao et al. \cite{twoattention} propose a two-level attention model, which combines two level attentions to select relevant proposals to the object and the discriminative parts. And Zhang et al. \cite{picking} incorporate deep convolutional filters for both part detection and description. The problem of fine-grained image classification is still far from solved. 
\begin{figure*}[!ht]
\begin{center}
\includegraphics[width=0.9\linewidth]{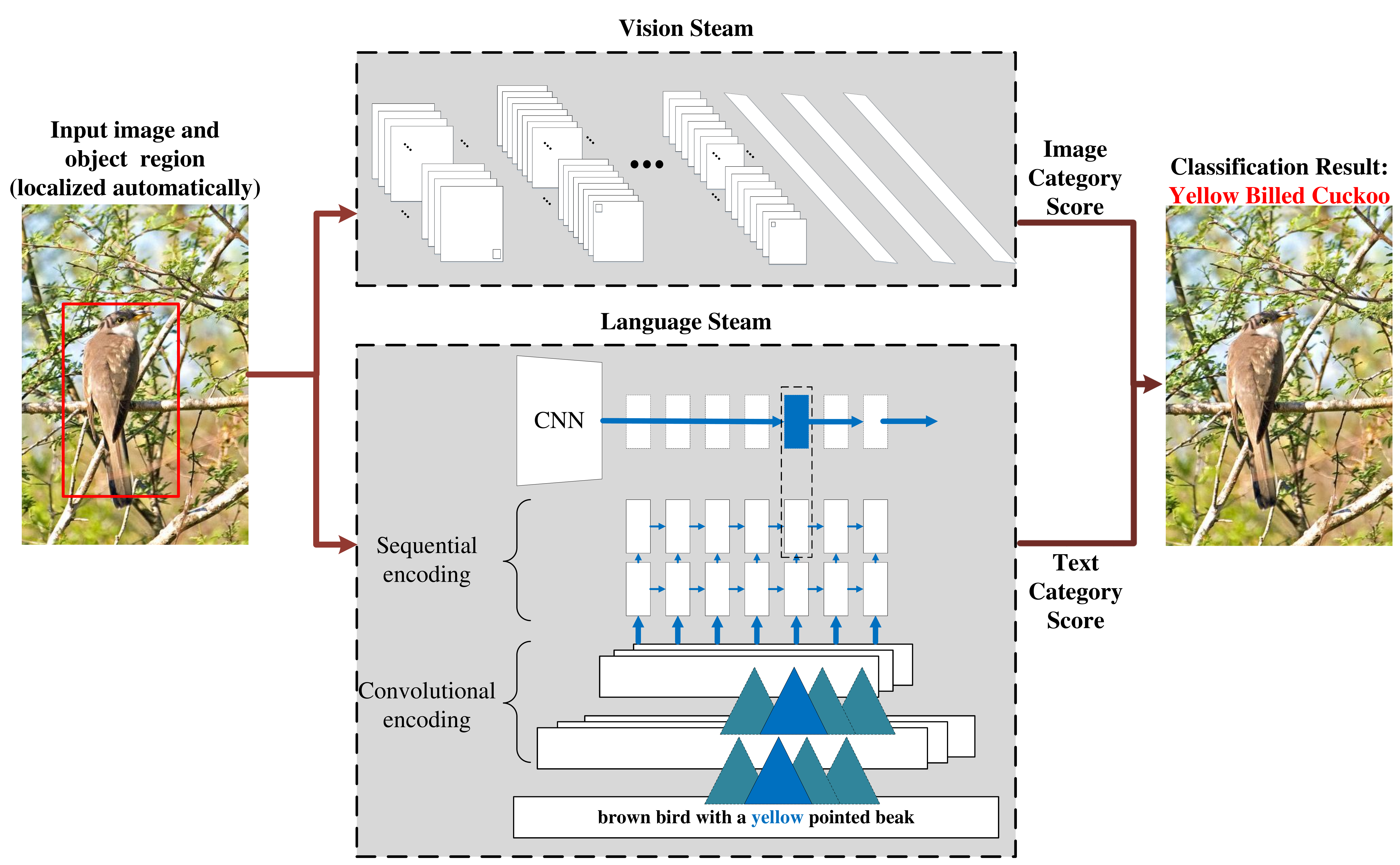}
\end{center}
   \caption{Overview of our CVL approach. The two-stream model conducts on the original images and their object localizations. One learns the deep representations directly from the vision information. The other learns the salient visual aspects for distinguishing sub-categories via jointly modeling vision and language. The classification results of the two streams are merged in later phase to combine the advantages of vision and language.}
\label{framework}
\end{figure*}
\subsection{Cross-modal analysis}
With the rapid growth of multimedia information, the cross-modal data, e.g. image, text, video and audio, has been the main form of the big data. Cross-modal data carries different kinds of information, which needs to be integrated to get comprehensive results in many real-world applications. How to learn multi-modal representation for cross-modal data is a fundamental research problem. 
A traditional representation method is the canonical correlation analysis (CCA) \cite{hotelling1936relations}, which learns a subspace to maximize the correlation among data of different media types, and is widely used for modeling multi-modal data \cite{bredin2007audio,hardoon2004canonical,klein2015associating}. Zhai et al. \cite{zhai2013heterogeneous} propose to learn projection functions by the metric learning, and this method is further improved as Joint Representation Learning (JRL) \cite{zhai2014learning} by adding other information such as semantic categories and semi-supervised information. Inspired by the progress of deep neural networks, some works have been focused on deep multi-modal representation learning.  Ngiam et al. \cite{ngiam2011multimodal} propose a multi-modal deep learning (MDL) method to combine the audio and video into an autoencoder, which improves the speech signal classification for noisy inputs as well as learns a shared representation across modalities. 
Recently, a surge of progress has been made in image and video captioning. LSTMs \cite{hochreiter1997long} are widely used in modeling captions at word level. Besides LSTMs, character-based convolutional networks \cite{zhang2015character} have been used for language modeling. In this paper, we apply the extension of Convolutional and Recurrent Networks (CNN-RNN) to learn a visual semantic embedding. 
In this paper, we bring the multi-modal representation learning into fine-grained image classification to boost the performance, and jointly modeling vision and language.

\section{Our CVL Approach}
Our method is based on a very simple intuition: natural language descriptions could point out the discriminative parts or characteristics from other sub-categories, and are complementary with vision information. Therefore, we propose a two-stream model combining vision and language for learning latent semantic representations, which takes the advantages of vision and language jointly, as shown in Figure. \ref{framework}. Since the object is crucial for fine-grained image classification, we take the original images and their object localizations as the inputs of the two-stream model. 

\subsection{Object localization}
\begin{figure}[t]
\begin{center}
\includegraphics[width=1\linewidth]{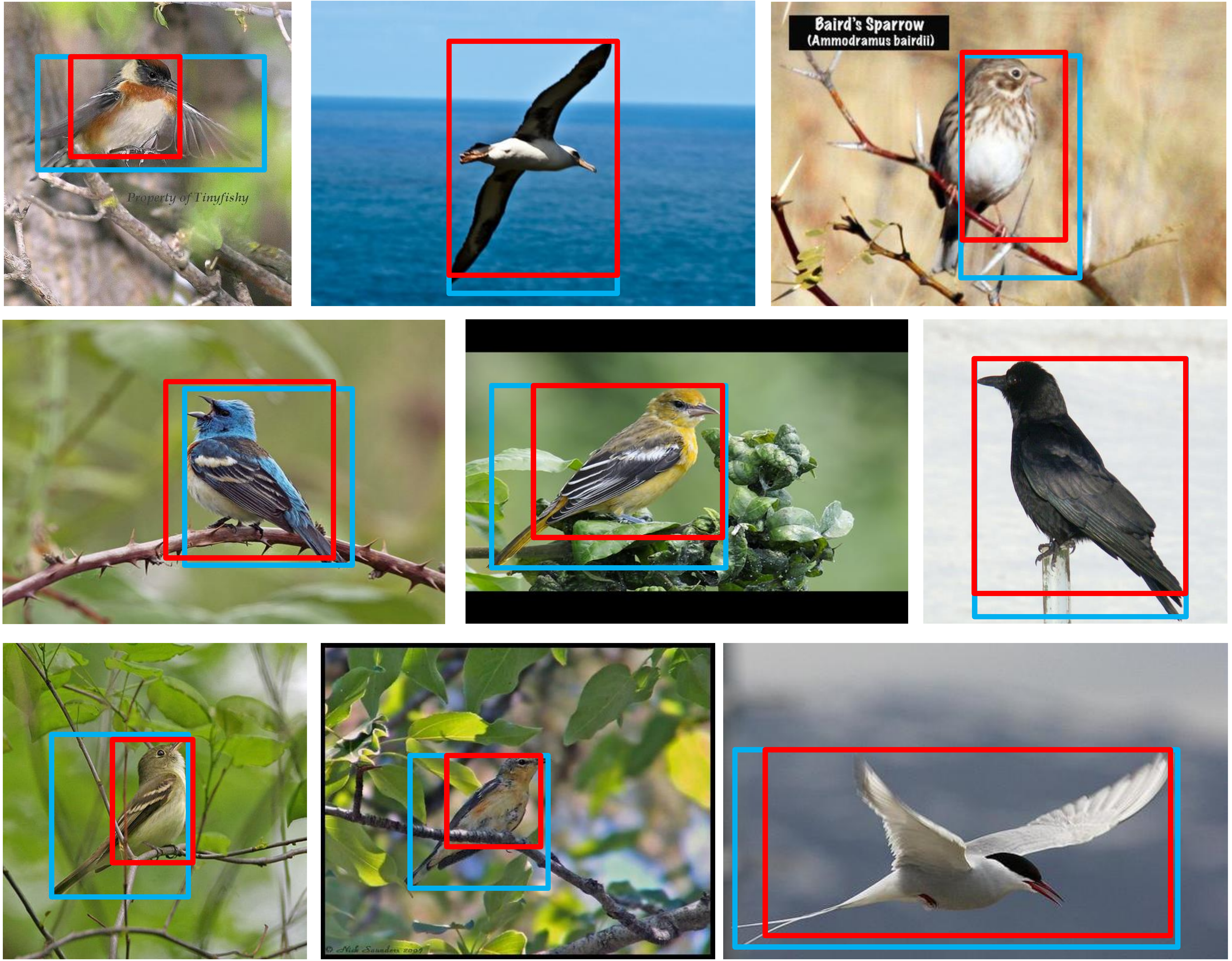}
\end{center}
   \caption{Sample object localization results in this paper. The blue rectangles indicate the ground truth object annotations, i.e. bounding boxes of objects, and the red rectangles indicate the object regions generated by jointly applying saliency extraction and co-segmentation.}
\label{object}
\end{figure}

In this paper, we apply an automatic object localization method based on saliency extraction and co-segmentation proposed in TSC \cite{spatialconstraints}, which allows to localize the object in a weakly-supervised manner that means neither object nor part annotations are used. Saliency extraction is to localize the object preliminarily with the saliency map generated by the CNN model. However, only through saliency extraction, the object region is not accurate enough so that co-segmentation is conducted to make the object region more accurate for fine-grained image classification. The sample object localization results are shown in Figure \ref{object}.

\begin{figure*}[t]
\begin{center}
\includegraphics[width=0.9\linewidth]{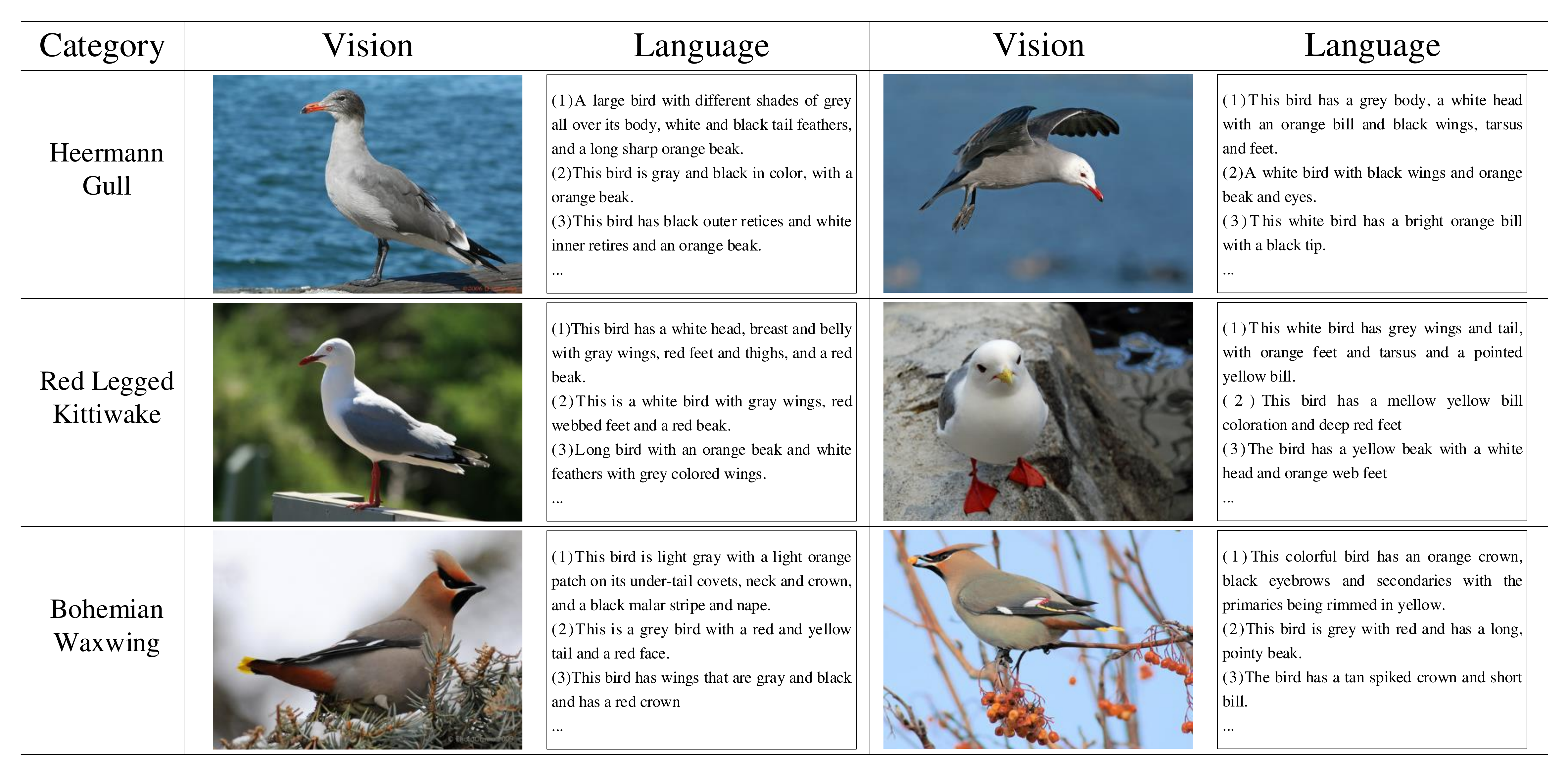}
\end{center}
   \caption{Sample natural language descriptions of CUB-200-2011.}
\label{language example}
\end{figure*}

\subsection{Jointly Model Vision and Language}
Considering that the two different descriptions of an image are complementary, i.e. visual information and natural language descriptions, we jointly model the two different forms of descriptions to learn deep representations for better classification accuracy.
\subsubsection{Vision Stream} 
A natural candidate for the visual classification function $f$ is a CNN model, which is consist of a hierarchy of convolutional and fully connected layers. We can benefit from model pre-training due to the additional training data. This has been proved by a large amount of recognition tasks, such as object detection, texture recognition and fine-grained image classification \cite{cimpoi2014describing,donahue2014decaf,rcnn,sharif2014cnn} etc. Therefore, we use a CNN model pre-trained on the ImageNet dataset \cite{imagenet} as the base model in our experiments. And then, we fine-tune the pre-trained CNN model on the fine-grained dataset. 
\par
Given an image $I$, its object region $b$ is generated at object localization stage, then the object region is clipped from the original image and saved as image $I'$. We take the original image $I$ and its object image $I'$ as the inputs of the CNN model to obtain the prediction, which is the result of the vision stream.
\subsubsection{Language Stream} 
\textbf{Deep Structured Joint Embedding} \ \ We apply the deep structured joint embedding method \cite{deeprepresentations}, because it can jointly embedding images and fine-grained visual descriptions (i.e. natural language descriptions for images). This method learns a compatibility function of image and text, which can be seen as an extension of the multimodal structured jointed embedding \cite{akata2015evaluation}. Instead of using a bilinear compatibility function, we use the inner product of features generated by deep neural encoders, and maximize the compatibility between a description and its matching image as well as minimize compatibility with images from other classes.

\par
Given data $D={(v_n,t_n,y_n), n=1, ..., N}$, in which $v \in V$ indicates visual information, $t \in T$ indicates text description and $y \in Y$ indicates the class label, then the image and text classifier functions $f_v : V \to Y$ and $f_t : T \to Y$ are learned by minimizing the empirical risk:

\begin{gather}
\frac{1}{N} \sum_{n=1}^N \Delta (y_n,f_v(v_n))+ \Delta (y_n,f_t(t_n))
\end{gather}
where $\Delta : y \times y \to \mathbb{R}$ is the 0-1 loss and
\begin{gather}
f_v(v)=arg \max_{y \in Y} \mathbb{E}_{t \thicksim T(y)}[F(v,t)] \\
f_t(t)=arg \max_{y \in Y} \mathbb{E}_{v \thicksim V(y)}[F(v,t)]
\end{gather}
We then define the compatibility function $F : V \times Y \to \mathbb{R}$ that uses features from the learnable encoder functions $\theta (v)$ for images and $\phi (t)$ for texts:
\begin{gather}
F(v,t)= \theta (v)^T \phi(t)
\end{gather}
We apply the GoogleNet \cite{googlenet} as the image encoder model, and Convolutional Recurrent Net (CNN-RNN) \cite{deeprepresentations} as the text encoder model which will be discussed in the next paragraph.
\par
\noindent{\textbf{Text encoder model}} \ \ We apply the CNN-RNN \cite{deeprepresentations} for learning the fine-grained visual descriptions. A mid-level temporal CNN hidden layer is at the bottom of CNN-RNN model, and a recurrent network is stacked on it. We extract the average hidden unit activation over the sequence as the text feature, as shown in equation \ref{encoder}. The resulting scoring function is defined as a linear accumulation of evidence for compatibility with the image which needs to be recognized.
\begin{gather}
\phi(t)= \frac{1}{L}\sum_{i=1}^L h_i
\label{encoder}
\end{gather}
where $h_i$ indicates the hidden activation vector for the $i$-th frame and $L$ indicates the sequence length.

\subsection{Final Prediction}
Given an Image $I$, its object bounding region is obtained automatically through localization method. The two-stream model conducts on the original images and their object localizations. The vision stream gives the prediction from the view of the image only, while the language stream gives the prediction via measuring the image and text description with the shared compatibility function. Finally, we fuse the prediction results of the two streams to utilize the advantages of the two via the follow equation:
\begin{gather}
f(I)= f_v(v) + \beta * f_t(t)
\end{gather}
where $f_v(v)$ and $f_t(t)$ are the image and text classifier functions as mentioned above, and $\beta$ is selected by the cross-validation method. In the experiments, we set $\beta$ as 3.

\begin{table*}[!ht]
   \centering
   \begin{tabular} {|c|c|c|c|c|c|c|}
      \hline
      \multirow {2}{*}{Method} & \multicolumn{2}{|c|}{Train Annotation} & \multicolumn{2}{|c|}{Test Annotation} & \multirow {2}{*}{Accuracy (\%)} \\ 
      \cline{2-5}
      &Bbox & Parts & Bbox & Parts &\\
      \hline
      \hline
      \textbf{Our CVL approach} & & & & & {\textbf{85.55}} \\
      \hline
      PD \cite{picking} & & & & & 84.54 \\
      Spatial Transformer \cite{jaderberg2015spatial} & & & & & 84.10 \\
      Bilinear-CNN \cite{lin2015bilinear} & & & & & 84.10 \\
      NAC \cite{simon2015neural} & & & & & 81.01  \\
      TL Atten \cite{twoattention} & & & & & 77.90  \\
      VGG-BGLm \cite{zhou2015fine} & & & & & 75.90  \\
      \hline
      PG Alignment \cite{krause2015fine} & $\surd$ & & $\surd$ & & 82.80 \\
      Triplet-A (64) \cite{cui2015fine} & $\surd$ & & $\surd$ & & 80.70\\
      VGG-BGLm \cite{zhou2015fine} & $\surd$ & & $\surd$ & & 80.40 \\
      Part-based R-CNN \cite{partrcnn} & $\surd$ & $\surd$ &  & & 73.50 \\
      \hline
      SPDA-CNN \cite{zhangspda} & $\surd$ & $\surd$ &$\surd$ &  & 85.14\\
      Part-based R-CNN \cite{partrcnn}& $\surd$ & $\surd$ & $\surd$ & $\surd$ & 76.37 \\
      POOF \cite{berg2013poof} & $\surd$ & $\surd$ & $\surd$ & $\surd$ & 73.30  \\
      GPP \cite{xie2013hierarchical} & $\surd$ & $\surd$ & $\surd$ & $\surd$ & 66.35  \\
      \hline
   \end{tabular}
   \caption{Comparisons with state-of-the-art methods on CUB-200-2011, sorted by amount of annotation used. ``Our CVL'' indicates our full method combining vision and language. ``Bbox'' indicates the object annotation (i.e. bounding box of object) provided by the dataset, and ``Parts'' indicates the parts annotations (i.e. parts locations). ``$\surd$'' indicates that one of bounding box and part locations is used in training or testing stage. Since the exact amount of annotation used varies from method to method, we defer to the original sources for details.}
   \label{cubresult}
\end{table*}

\section{Experiments}
This section presents the evaluations and analyses of our CVL approach on the challenging fine-grained image classification benchmark CUB-200-2011 \cite{cub2011}. It contains 11,788 images of 200 types of birds, 5,994 for training and 5,794 for testing. Every image has detailed annotations: 15 part locations, 312 binary attributes and 1 bounding box. Scott Reed et al. \cite{deeprepresentations} expand the CUB-200-2011 dataset by collecting fine-grained visual descriptions. Ten single-sentence visual descriptions are collected for each image, as shown in Figure \ref{language example}. The fine-grained visual descriptions are collected through the Amazon Mechanical Turk (AMT) platform, and are at least 10 words, without any information of species, background and actions. 

\subsection{Implementation Details}
\noindent{\textbf{Vision Stream}} \ \ In our experiments, we apply the widely used model of VGGNet \cite{vgg} as the vision stream model. The reason of choosing VGGNet is for fair comparison with state-of-the-art methods. It is important to note that the model used in our proposed method can be replaced with any CNN model. The model is pre-trained on ImageNet dataset, and then fine-tuned on the CUB-200-2011 dataset. In the fine-tuning step, we follow the strategy of TL Atten \cite{twoattention}. First, we apply the selective search \cite{uijlings2013selective} to generate patches for each image. Then the pre-trained CNN model on ImageNet dataset is used as a filter net for selecting the patches relevant to the object. With the selected patches, we fine-tune the pre-trained model.

\par
\noindent{\textbf{Language Stream}} \ \ In our experiments, we apply the GoogleNet \cite{googlenet} with batch normalization \cite{ioffe2015batch} as image encoder and CNN-RNN \cite{deeprepresentations} as text encoder. For image encoder, we take the strategy used in vision stream for better accuracy. And for text encoder, the CNN input size (sequence length) is set to 201 for character-level model. We keep the image encoder fixed, and used RMSprop with base learning rate 0.0007 and minibatch size 40. All the configurations and source code \footnote{https://github.com/reedscot/cvpr2016} used for training and testing follow the work of Scott Reed et al. \cite{deeprepresentations}.

\subsection{Comparisons with state-of-the-art methods}
For comparison purpose, we adopt 12 state-of-the-art fine-grained image classification methods. Table \ref{cubresult} shows the comparison results on CUB-200-2011. Bounding box and part annotations used in the methods are listed for fair comparison. Early works \cite{berg2013poof,xie2013hierarchical} choose SIFT \cite{lowe2004distinctive} as features, and the performance is limited. When applying CNN model, our CVL approach is the best. In our experiments, both object and part annotations are not used, due to labeling is heavily labor consuming. Compared with the methods \cite{jaderberg2015spatial,lin2015bilinear,simon2015neural,twoattention,picking} which do not use object and part annotations, our CVL approach obtains a 1.01\% higher accuracy than the best performing result of PD \cite{picking}. Moreover, our CVL approach outperforms methods which use object annotation \cite{krause2015fine} (82.50\%) or even part annotations \cite{zhangspda,zhang2014part} (85.14\%, 76.37\%). 
It proves the effectiveness of our CVL approach, which jointly integrates the vision and language streams to exploit the correlation between visual feature and nature language descriptions and enhance their complementarity.

\begin{figure*}[t]
\begin{center}
\includegraphics[width=0.8\linewidth]{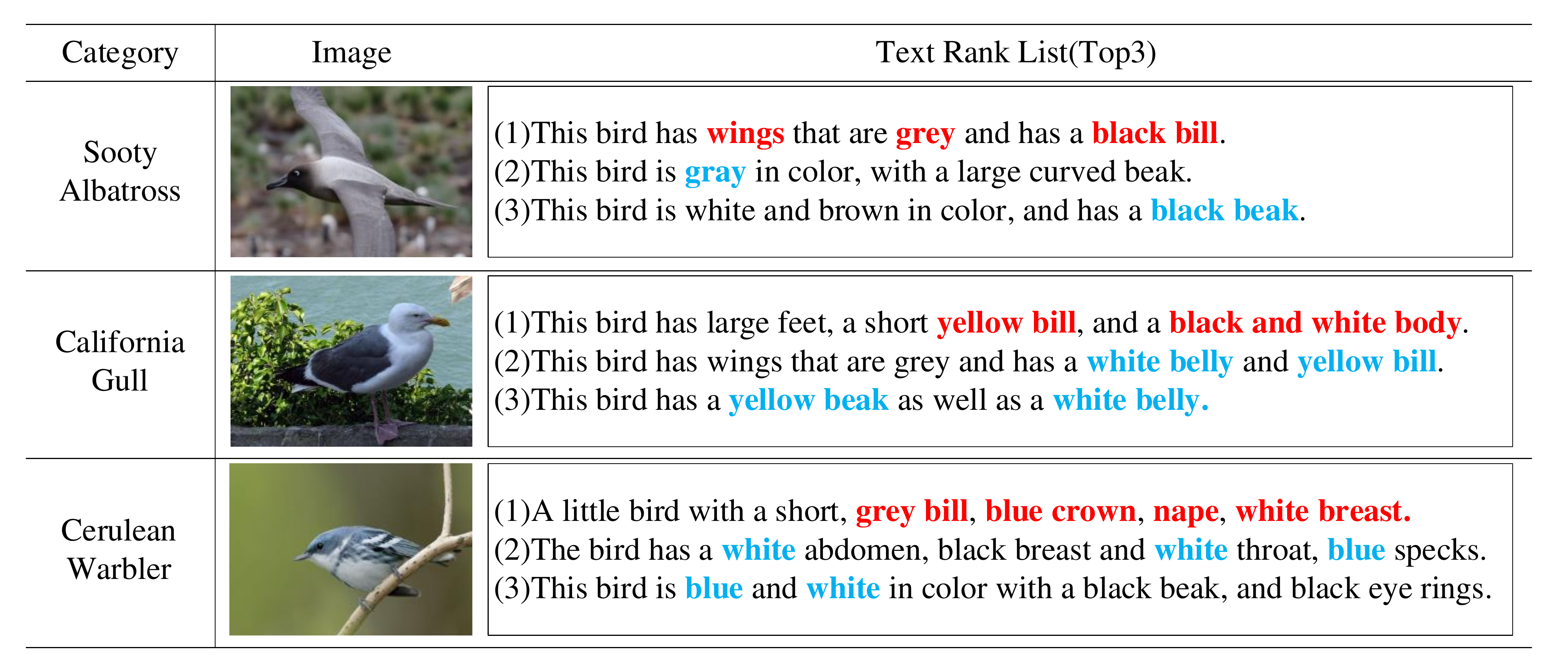}
\end{center}
   \caption{Some results of the language stream. The red words are the important visual descriptions for distinguishing sub-categories, and the blue ones are the visual descriptions of the easily confused sub-categories.}
\label{visonandlanguage}
\end{figure*}

\subsection{Performances of components in our CVL approach}
\subsubsection{Effectivenesses of vision stream and language
stream}
We perform detailed analyses by comparing different variants of our CVL approach. ``Language-stream'' refers to the classification result of the language stream, ``Vision-stream'' refers to the classification result of the vision stream, ``CVL'' refers to our CVL approach combining vision and language, and ``Original'' refers to the classification result that only use the original image for prediction. From Table \ref{variants}, we can observe that:

\begin{table}
   \begin{center}
   \begin{tabular} {|c|c|}
      \hline
      Method & Accuracy (\%)\\ 
      \hline
      \hline
      {\begin{tabular}{c} \textbf{our CVL approach} \\ \textbf{(Language-stream+Vision-stream)} \end{tabular} }& {\textbf{85.55}} \\ 
      \hline
      Language-stream & 81.81 \\
      \hline
      Vision-stream & 82.98 \\
      \hline
      Original & 76.17 \\
      \hline
   \end{tabular}
   \end{center}
   \caption{Effects of different variants of our method on CUB-200-2011. ``Language-stream'' refers to the classification result of the language stream, ``Vision-stream'' refers to the classification result of the vision stream, ``CVL'' refers to our CVL approach combining vision and language, and ``Original'' refers to the classification result that only use the original image for prediction.}
   \label{variants}
\end{table}
\begin{itemize}
\item
Two-stream model combining vision and language boosts the performance significantly. CVL brings about a nearly 10\% (76.17\% $\to$ 85.55\%) improvement compared with the ``Original''.
\item
The classification result of the language stream is promising. From the first line of each row in Figure \ref{visonandlanguage}, we can find that the text description with the highest score always points out the discriminative parts or characteristics. As shown in Figure \ref{visonandlanguage} , the red words are the important visual descriptions for distinguishing sub-categories, and the blue ones are the visual descriptions of the easily confused sub-categories.

\item
Combining vision and language can achieve more accurate result than only one stream (85.55\% vs. 81.81\% and 82.98\%), which demonstrates that visual information and text descriptions are complementary in fine-grained image classification. The two streams have the different but complementary focuses. (1) The vision stream localizes the object region of image and extracts the visual features from the original pixels through the CNN model, which focuses on the location of the discriminative region and the texture, color or even the semantic parts we called. However, we do not know or learn correctly which parts or features are the most discriminative representations from other sub-categories. (2) The language stream learns correlation between the nature language descriptions and the visual features to exploit the attributes of the discriminative regions for distinguishing sub-categories. The natural language descriptions directly point out key parts or features distinguished from other sub-categories, e.g. Cerulean Warbler has blue crown, back and white breast, while Sooty Albatross has grey wings and black bill.
\end{itemize}

\subsubsection{Effect of fine-tuning and object localization}
There are two differences from the work \cite{deeprepresentations}: (1) instead of directly using the GoogleNet, we first fine-tune it on the CUB-200-2011, and (2) extract the features of the original image and its object region for each image. We find that both (1) and (2) are important for not only fine-grained image classification but also zero-shot recognition, as shown in Table \ref{fine} and Table \ref{zero} respectively. It also proves the effectiveness of object localization in vision stream of our CVL approach, which focuses on the discriminative region of the image and eliminate the side effect of the background noise.

\begin{table}
   \begin{center}
   \begin{tabular} {|c|c|}
      \hline
      Method & Accuracy (\%)\\ 
      \hline
      \hline
      \textbf{Language+ft+box} & {\textbf{81.81}} \\ 
      \hline
      Language+ft & 77.80 \\
      \hline
      Language & 50.54 \\
      \hline
   \end{tabular}
   \end{center}
   \caption{Effect of fine-tuning and object localization for fine-grained image classification. ``ft'' indicates fine-tuning is applied, and ``box'' indicates object localization is applied.}
   \label{fine}
\end{table}

\begin{table}
   \begin{center}
   \begin{tabular} {|c|c|}
      \hline
      Method & Top-1 Accuracy (\%)\\ 
      \hline
      \hline
      \textbf{DS-SJE+ft+box} & {\textbf{65.1}} \\ 
      \hline
      DS-SJE+ft & 60.0 \\
      \hline
      DS-SJE \cite{deeprepresentations} & 54.0 \\
      \hline
   \end{tabular}
   \end{center}
   \caption{Effect of fine-tuning and object localization for zero-shot recognition. ``ft'' indicates fine-tuning is applied, and ``box'' indicates object localization is applied.}
   \label{zero}
\end{table}

\section{Conclusions}
In this paper, the CVL approach has been proposed, which jointly models vision and language for learning latent semantic representations. The vision stream learns deep representations from the original visual information via deep convolutional neural network. The language stream utilizes the natural language descriptions which could point out the discriminative parts or characteristics for each image, and provides a flexible and compact way of encoding the salient visual aspects for distinguishing sub-categories. Since the two streams are complementary, combining the two streams can further achieves better classification accuracy. Experimental results on CUB-200-2011 dataset demonstrate the superiority of our method compared with state-of-the-art methods. The results are promising, and point out a few future directions. First, combining vision and language can boosts the classification accuracy, but the two streams are trained respectively, we will focus on the work of training the two streams end-to-end. Second, from Table \ref{fine} we can find that small improvement on the original language stream boosts the performance a lot. And nowadays there are a lot of works focusing on how to relate images to natural language descriptions. So improving the performance of the language stream will be significantly helpful to the fine-grained image classification.
\section{Acknowledgments}
This work was supported by National Natural Science Foundation of China under Grants 61371128 and 61532005.

\balance 
{\small
\bibliographystyle{ieee}
\bibliography{reference}
}
{}
\end{document}